\documentclass[sigconf, nonacm]{acmart}

\usepackage{times}  
\usepackage{helvet}  
\usepackage{courier}  
\usepackage{graphicx} 
\usepackage{natbib}  
\usepackage{caption} 
\usepackage{algorithmic}

\usepackage[utf8]{inputenc} 
\usepackage[T1]{fontenc}    
\usepackage{url}            
\usepackage{booktabs}       
\usepackage{amsfonts}       
\usepackage{nicefrac}       
\usepackage{microtype}      
\usepackage{xcolor}         
\usepackage{amsmath}

\usepackage{amsmath}
\usepackage{float}
\usepackage{multirow}
\usepackage{booktabs}
\usepackage{cite}
\usepackage[ruled,vlined]{algorithm2e}
\usepackage{pifont}
\usepackage{diagbox}
\usepackage{boldline}
\usepackage{array}
\usepackage{arydshln}

\usepackage{makecell}

\usepackage{newfloat}
\usepackage{listings}

\AtBeginDocument{%
  }

\setcopyright{acmlicensed}

\makeatletter
\def\@authorfont{\Large\bfseries}      
\def\@affiliationfont{\large\normalfont} 
\makeatother

\begin{document}

\title{VAGPO: Vision-augmented Asymmetric Group Preference Optimization for Graph Routing Problems}

\author{Shiyan Liu}
\affiliation{%
  \institution{Huazhong University of Science and Technology}
  \city{Wuhan}
  \country{China}}
\email{shyl@hust.edu.cn}

\author{Bohan Tan}
\affiliation{%
  \institution{Huazhong University of Science and Technology}
  \city{Wuhan}
  \country{China}}
\email{bohant@hust.edu.cn}

\author{Zhiguang Cao}
\affiliation{%
  \institution{Singapore Management University}
  \city{}
  \country{Singapore}}
\email{zhiguangcao@outlook.com}

\author{Yan Jin}
\affiliation{%
  \institution{Huazhong University of Science and Technology}
  \city{Wuhan}
  \country{China}}
\email{jinyan@mail.hust.edu.cn}

\settopmatter{printacmref=false}

\begin{abstract}
Graph routing problems play a vital role in web-related networks, where finding optimal paths across graphs is essential for efficient data transmission and content delivery. Classic routing formulations such as the Traveling Salesman Problem (TSP) and the Capacitated Vehicle Routing Problem (CVRP) represent fundamental graph optimization challenges. Recent data-driven optimization methods have made significant progress, yet they often face limitations in training efficiency and generalization to large-scale instances. In this paper, we propose a novel Vision-augmented Asymmetric Group Preference Optimization (VAGPO) approach. By leveraging ResNet-based visual encoding and Transformer-based sequential modeling, VAGPO captures both spatial structure and temporal dependencies. Furthermore, we introduce an asymmetric group preference optimization strategy that significantly accelerates convergence compared to commonly used policy gradient methods. Experimental results on generated TSP and CVRP instances, as well as real-world datasets, demonstrate that the proposed VAGPO approach achieves highly competitive solution quality. Additionally, VAGPO exhibits strong generalization to larger instances (up to 1000 nodes) without re-training, highlighting its effectiveness in both learning efficiency and scalability.
\end{abstract}



\maketitle

\section{INTRODUCTION}

Graph routing optimization is fundamental to modern communication networks, where the goal is to determine the most efficient paths for data transmission given the network topology. Effective routing not only minimizes latency and maximizes throughput but also enhances overall resource utilization, scalability, and resilience across diverse web-scale applications such as the Internet, cloud computing, and content delivery networks \citep{jiang2024graph}. Among the broad family of graph optimization problems, the Traveling Salesman Problem (TSP) and the Capacitated Vehicle Routing Problem (CVRP) stand out as canonical formulations with direct implications for network design and optimization \citep{baldacci2010some}. These problems have been extensively investigated in both operations research and artificial intelligence communities due to their rich combinatorial structure and practical relevance. However, as NP-hard problems, their computational complexity increases exponentially with problem size, rendering traditional exact solvers such as Concorde \citep{applegate2006concorde}, infeasible for large-scale instances. Consequently, a wide range of heuristic and metaheuristic algorithms have been proposed to efficiently generate high-quality, near-optimal solutions suitable for real-world deployment.


In recent years, a soaring number of studies of data-driven optimization algorithms based on deep learning with supervised and reinforcement learning have been proposed to solve graph routing problems. Most of these approaches treat the problems as a sequence generation task using the sequence-to-sequence architectures. Broadly, they can be categorized into constructive and iterative approaches. Constructive approaches incrementally generate a complete solution by adding one node at a time to a partial route. These approaches are highly efficient during inference, making them well-suited for real-time applications, such as Attention Model (AM) \citep{kool2018attention}, Pointerformer \citep{jin2023pointerformer}, Policy Optimization with Multiple Optima (POMO) \citep{kwon2020pomo}, and its Preference Optimization (PO)-based version POMO+PO \citep{pan2025preference}. In contrast, iterative approaches start from an initial solution and iteratively improve it through a neighborhood search based on certain local operators until a termination condition is met. These approaches, such as \citet{wu2021learning}, tend to be more computationally intensive during training and inference, but they often yield superior solution quality over time.

While most data-driven optimization algorithms focus on sequential representations, relatively few studies have explored structured representations that explicitly capture the spatial and topological properties of the graph routing problems. Graph Neural Networks (GNNs) \citep{scarselli2008graph} have been introduced to exploit the inherent graph structure of these problems, enabling the modeling of complex node relationships and topological features that are often challenging to capture with purely sequential models \citep{prates2019learning, xing2020graph}. Most recently, vision-based approaches have emerged, where graph instances are encoded into image-like representations to exploit spatial patterns. These approaches leverage the observation that graph problems exhibit intrinsic spatial regularities, which can be effectively modeled using Convolutional Neural Networks (CNNs) \citep{lecun1989backpropagation}. For example, H-TSP \citep{pan2023h} employs CNNs to process TSP instances as images, extracting spatial features that guide the optimization process.

In this paper, we propose a Vision-augmented Asymmetric Group Preference Optimization (VAGPO) approach for solving the graph routing problems in a constructive manner. Built upon the classical encoder-decoder architecture \citep{vaswani2017attention}, VAGPO incorporates a vision-based representation by transforming graph instances into image-like formats and utilizing a ResNet architecture \citep{he2016deep} to extract both local and global spatial features. These features complement the sequential modeling capabilities of Transformer-based architectures, enabling improved spatial-temporal reasoning for trajectory generation.

The main contributions of our work are summarized as follows: 
\begin{itemize}
\item We propose a cross-modal feature fusion mechanism that effectively integrates sequential and visual information, enabling the model to jointly reason over temporal decision-making and spatial patterns by leveraging both local and global features.
\item We introduce a reinforcement learning strategy, Asymmetric Group Preference Optimization (AGPO), which generalizes preference optimization to the multi-start node setting. By incorporating asymmetric optimization parameters and group-wise trajectory evaluations, AGPO improves training stability and convergence speed.
\item The proposed VAGPO achieves highly competitive performance across all generated TSP and CVRP instances and real-world instances, while requiring substantially fewer training epochs than existing learning-based methods. Furthermore, it generalizes well to instances with varied distributions and scales without re-training.
\end{itemize}

\section{RELATED WORK}
Traditional approaches can be grouped into exact and heuristic algorithms. Concorde \citep{applegate2006concorde} is one of the most well-known exact solvers, which formulates the problem to mixed integer programming and solves it by a branch-and-cut framework. LKH3 \citep{helsgaun2017extension} is a state-of-the-art (SOTA) heuristic using neighborhood search and a population-based framework to find high-quality solutions.

With the rapid progress of deep learning, data-driven optimization approaches have gained increasing attention. They are generally classified into constructive and iterative algorithms. Constructive algorithms generate solutions incrementally by sequentially selecting nodes. Pointer Network \citep{vinyals2015pointer} was the first to apply deep learning to graph problems. Building on this, \citet{bello2016neural} introduced a reinforcement learning (RL) framework that eliminates the need for optimal labels. \citet{nazari2018reinforcement} improved representation by mapping node features into a shared high-dimensional space. Inspired by Transformers \citep{vaswani2017attention}, \citet{kool2018attention} proposed the AM replacing RNNs in Seq2Seq structures with attention modules. Further developments include Multi-Decoder AM \citep{xin2021multi} and POMO \citep{kwon2020pomo}, which leverage problem symmetries to improve performance. Sym-NCO \citep{kim2022sym} generalizes symmetry-aware learning across broader tasks, and POMO+PO \citep{pan2025preference} further advances the field by integrating preference-based learning into SOTA models. Pointerformer \citep{jin2023pointerformer} improves the Transformer backbone, and Poppy \citep{grinsztajn2024winner} introduces diverse policy populations without relying on explicit sampling strategies.

In contrast, iterative approaches refine an initial solution through neighborhood search, combining traditional solvers with learning-based guidance. \citet{lu2019learning} integrated RL with LKH to control handcrafted improvement rules. \citet{d2020learning} and \citet{sui2021learning} applied RL to guide 2-opt and 3-opt operators. NLNS \citep{hottung2020neural} merged RL with large neighborhood search \citep{shaw1998using}. NeuroLKH \citep{xin2021neurolkh} used a supervised sparse graph network to generate candidate edge sets. SGBS \citep{choo2022simulation} combined neural policy networks with simulation-based tree search, while Neural $k$-Opt \citep{ma2024learning} employed a dual-stream decoder and action factorization for adaptive exchanges.

Recently, diffusion models have emerged as promising alternatives. DIFUSCO \citep{sun2023difusco} and T2T \citep{li2023t2t} apply diffusion-based generative techniques to solve graph optimization problems, highlighting the potential of continuous generative models in discrete domains.

To overcome the limitations of sequential representations, researchers have explored structured information to better capture geometric and topological properties, graph-based models have shown strong potential. \citet{prates2019learning} demonstrated that GNNs \citep{scarselli2008graph} can solve NP-complete problems by integrating symbolic and numerical reasoning. GREAT \citep{xing2020graph} further improved efficiency through edge sparsification while preserving key structural information.

Vision-based representations also offer a promising alternative. H-TSP \citep{pan2023h} uses a hierarchical framework where CNNs \citep{lecun1989backpropagation} encode TSP instances as images, capturing spatial relationships between nodes. These visual features reveal patterns often missed by sequence models. More recently, hybrid architectures such as the lightweight CNN-Transformer model \citep{jung2024lightweight} have sought to combine convolutional spatial encoding with attention-based sequential reasoning, achieving a more balanced trade-off between spatial perception and temporal decision-making. However, most vision-based methods still lack fully effective integration with sequential processing, limiting their ability to jointly reason over spatial and temporal dependencies—an essential capability for high-quality graph solutions.

\section{PROBLEM DEFINITION}
\textbf{Traveling Salesman Problem (TSP)} \quad Let $G = (V, E)$ denote an undirected graph, where $V$ represents the set of $N$ nodes and $E$ represents the set of edges connecting these nodes. For each edge connecting nodes $i$ and $j$, we define $cost(i, j)$ as the Euclidean distance between them. A special node $v_d \in V$ is designated as the depot where the route begins and ends. A feasible solution to a TSP instance is a Hamiltonian cycle that visits each node exactly once. Our objective is to minimize the total cost of the solution route $\tau$, which can be calculated by Eq. (\ref{eq_TSP}), where $v_i$ represents the $i$-th node in the route. All node coordinates are normalized to lie within $[0, 1]^2$.
\begin{equation} \label{eq_TSP}
 L(\tau) = \sum_{i=1}^{N-1} cost(v_i,v_{i+1}) + cost(v_N, v_1)
\end{equation}
\textbf{Capacitated Vehicle Routing Problem (CVRP)} \quad Let $G = (V, E)$ denote an undirected graph with $N$ nodes and associated edges. Each edge has a cost defined as the Euclidean distance between its endpoints. Node $v_d \in V$ serves as the depot. Each non-depot node, called the customer node, has an associated demand $d$. Each vehicle has a maximum capacity $Q$.

\begin{figure*}[!htb]
    \centering
    \includegraphics[width=\linewidth]{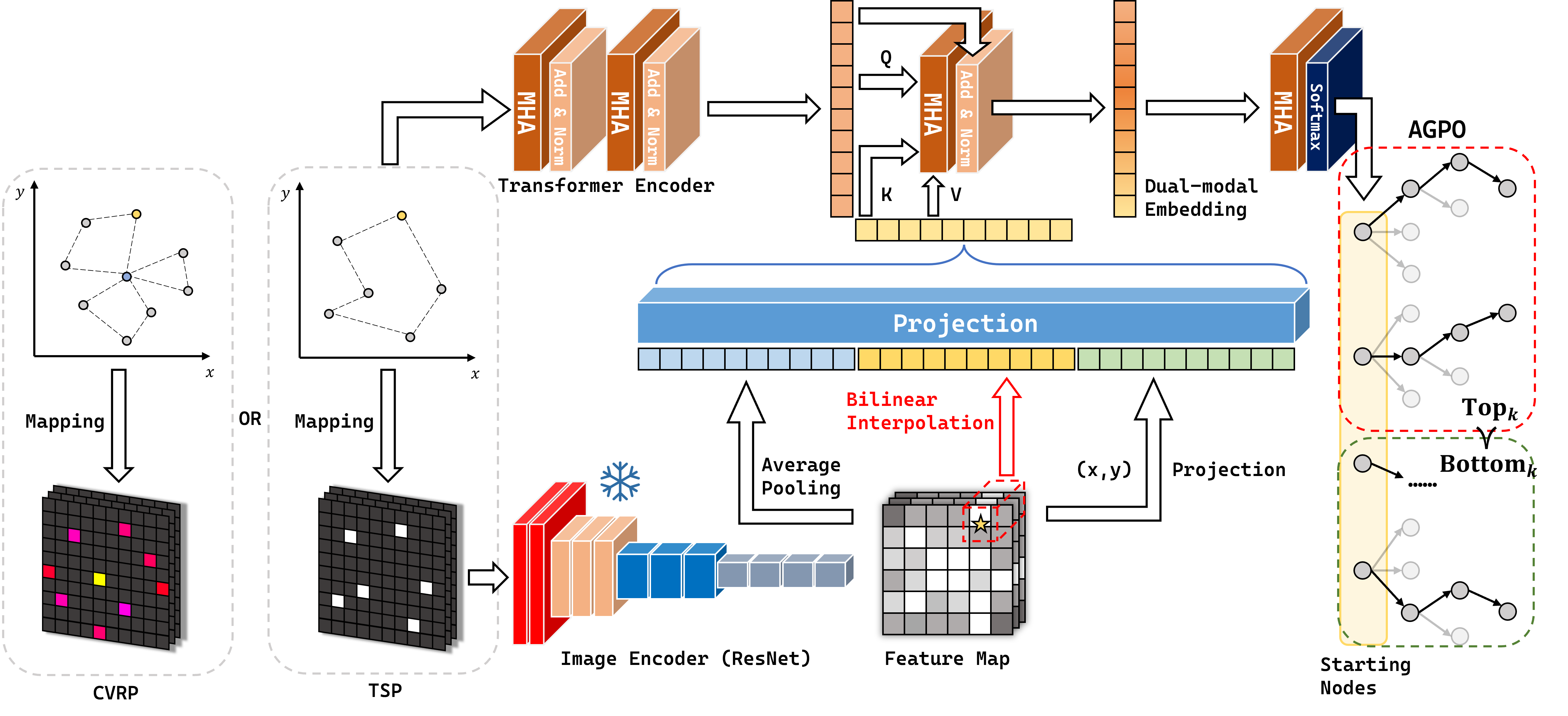}
    \caption{The overall framework of the proposed VAGPO approach}
    \label{fig:methods}
\end{figure*}

A feasible CVRP solution consists of a set of routes where: (1) each route starts and ends at the depot, (2) each non-depot node is visited exactly once by exactly one route, and (3) the total demand served by each route does not exceed $Q$. The objective is to minimize the total cost across all routes, can be calculated as in Eq. (\ref{eq_CVRP}), where $v^{k}_{i}$ is the $i$-th node in the $k$-th route, $m_k$ is the number of nodes in that route, and $v^{k}_{1} = v^{k}_{m_k} = v_d$. All node coordinates are normalized to lie within $[0, 1]^2$.
\begin{equation} \label{eq_CVRP}
L(\tau) = \sum_{k=1}^{K} \sum_{i=1}^{m_k-1} cost(v^{k}_{i}, v^{k}_{i+1})
\end{equation}

\section{THE PROPOSED VAGPO APPROACH}

\label{sec-method}
The proposed VAGPO approach is a constructive algorithm based on the Transformer architecture \citep{vaswani2017attention}, which combines a cross-modal encoder and an auto-regressive decoder to generate high-quality solutions for graph routing problems. The general framework of VAGPO is illustrated in Figure \ref{fig:methods}. 

In principle, VAGPO employs a vision representation that transforms node coordinates into an image-like format and uses the ResNet \citep{he2016deep} architecture to extract local and global visual features, effectively capturing geometric relationships among nodes. These visual features are then integrated with sequential information via a cross-modal feature fusion mechanism, which combines the spatial representations extracted by CNNs \citep{lecun1989backpropagation} with the sequence modeling capabilities of a Transformer encoder. The resulting node embeddings are passed to an auto-regressive decoder, which constructs the graph solution by selecting one node at a time based on the learned probability distribution. To improve training efficiency, VAGPO is trained using a reinforcement learning strategy AGPO, which extends preference-based learning to a multi-start setting. This training paradigm allows diverse trajectory evaluation and policy refinement.

\subsection{Vision Representation}
\label{vbpr}
To better capture the spatial structure inherent in TSP and CVRP, VAGPO begins by transforming each problem instance into an image-like representation. This transformation encodes the spatial layout of nodes in a format amenable to convolutional processing. To extract meaningful spatial features, we adopt a CNN-based encoder, specifically leveraging the ResNet architecture. This encoder is designed to learn both global contextual features of the instance and local geometric patterns surrounding each node, thereby enriching the encoder's embedding of spatial dependencies crucial for effective graph generation.

For the TSP, we convert the set of node coordinates $\{(x_i, y_i)\}_{i=1}^N$, where $N$ is the number of nodes and $(x_i, y_i)$ denotes the normalized 2D coordinate of node $v_i$, into a fixed-size grayscale image of $224 \times 224$. Each node $v_i$ is mapped to its corresponding pixel location $(x_i, y_i)$, where the pixel value is set to $1$ at the node's location and $0$ elsewhere. The resulting image is then duplicated across three channels to form a $224 \times 224 \times 3$ image. Given that $N$ is much smaller than the number of pixels $M$, the probability that at most one pixel contains multiple nodes can be approximated using the binomial formula:
\begin{equation} \label{eq:at_most_one_collision}
    P \approx (1 - p)^{M} + M p (1-p)^{M-1},
\end{equation}
where $p$ is the probability that at least one pixel contains multiple nodes:
\begin{equation} \label{eq:collision_probability}
    p = 1 - e^{-\frac{N}{M}} - \frac{N}{M} e^{-\frac{N}{M}}
\end{equation}
For $N=100$ and $M=224 \times 224$, this probability is approximately $99\%$, indicating that the impact of node collisions is negligible in practice.

For the CVRP, we convert the nodes into a three-channel image to distinguish different node types and encode demand information, where $d_j$ denotes the normalized demand of customer node $v_j$.
Specifically, let $P_{(x, y)} = [c_1, c_2, c_3]$ denote the value of the three channels at pixel $(x, y)$, where $(x, y)$ is a pixel position in the image.
The channel values are defined as:
\begin{equation} \label{eq_Pxy}
    P_{(x, y)} = 
    \begin{cases}
        [1, 1, 0], & \text{if depot $v_d$ at $(x, y)$} \\
        [1, 0, d_j], & \text{if customer $v_j \neq v_d$ at $(x, y)$} \\
        [0, 0, 0], & \text{otherwise}
    \end{cases}
\end{equation}


In this way, we have shifted the problem representation from numerical to visual. After this, we use a pretrained and frozen ResNet-18 as the image encoder to extract the feature map $G \in \mathbb{R}^{B \times C \times H \times W}$, where $B$ is the batch size, $C$ is the number of channels, and $H$, $W$ are the height and width of the feature map, respectively. To fully utilize the global and local features, we first perform adaptive average pooling on the feature map to obtain $G$, and then derive the feature embedding $F_i$ for node $v_i$ by bilinear interpolation from the same feature map according to its normalized coordinates $(x_i, y_i)$. Finally, the embedding $F^{\prime}_{i}$ of $v_i$ consists of a combination of $G$, $F_i$, and the positional encoding $P_i$ of $v_i$, as shown in Eq.(\ref{eq_encoding1}) and Eq.(\ref{eq_encoding2}), where
\begin{equation} \label{eq_encoding1}
    P_i = \text{Linear}([x_i, y_i])
\end{equation}
\begin{equation} \label{eq_encoding2}
    F^{\prime}_{i} = \text{Linear}(G \oplus F_i \oplus P_i)
\end{equation}

After projection, the resulting $F^{\prime}_{i} \in \mathbb{R}^{B \times D}$, where $D$ is the embedding dimension, possesses the same shape as the output of the Transformer encoder, thereby facilitating seamless integration in the subsequent cross-modal feature fusion stage. Despite the inevitable information loss introduced by visual encoding, its inherent locality serves as a valuable complement to the global modeling capacity of sequential information. 

\subsection{Cross-modal Feature Fusion}
To effectively integrate both spatial and sequential information in graph problems, we design a cross-modal feature fusion module within the encoder-decoder framework. While the Transformer architecture remains the backbone for modeling sequential dependencies, we incorporate visual features extracted from the image-based representation to enhance spatial awareness. By aligning the sequential and visual feature tensors into a common embedding space, the fusion mechanism enables the model to jointly reason over spatial structures and graph sequences, unlocking complementary strengths from both modalities.

Since both the sequential and visual features are preprocessed into tensors of the same dimensionality, the fusion network, mainly populated by a multi-head attention (MHA) layer and a feed forward (FF) layer, is adopted along with residual connections. By enabling each node in the sequential feature space to attend to multiple subspaces of the visual modality, the multi-head attention mechanism allows the model to capture diverse and complementary cross-modal information, thus enhancing the effectiveness of feature fusion. Let $T \in \mathbb{R}^{B \times N \times D}$ denote the sequential features obtained from the Transformer encoder and $V \in \mathbb{R}^{B \times N \times D}$ denote the visual features extracted from the ResNet encoder. The fused features $F$ are computed as in Eq. (\ref{eq_fusion}).
\begin{equation} \label{eq_fusion}
    F = T + \alpha_1 \cdot \text{MHA}_{T \rightarrow V}(T, V) + \alpha_2 \cdot \text{FF}(T + \alpha_1 \cdot \text{MHA}_{T \rightarrow V}(T, V))
\end{equation}
Here, $\alpha_1, \alpha_2 \in \mathbb{R}$ are learnable scaling factors, and $\text{FF}(\cdot)$ denotes a feed-forward network. The fusion proceeds in two stages: (1) $\text{MHA}_{T \rightarrow V}(T, V)$ enriches $T$ with visual information, scaled by $\alpha_1$ and added via residual connection; (2) the result is refined by $\text{FF}(\cdot)$, scaled by $\alpha_2$, and combined with $T$.

Specifically, $\text{MHA}_{T \rightarrow V}$ refers to a multi-head attention mechanism where sequential features $T$ act as queries ($Q$) and visual features $V$ serve as keys and values ($K$, $V$). This design leverages the complementary nature of the two modalities: $T$ captures global dependencies while $V$ provides local geometric patterns. Using $T$ as queries allows the model to selectively retrieve relevant spatial information from $V$ based on sequential context. For a single attention head:
\begin{equation} \label{Eq_attention}
    \text{Attention}(T, V) = \text{softmax}\left(\frac{(W_q T)(W_k V)^T}{\sqrt{D}} \cdot \beta\right) (W_v V)
\end{equation}
where $W_q, W_k, W_v \in \mathbb{R}^{D \times D}$ are learnable projection matrices and $\beta$ is a learnable scaling factor. The mechanism computes similarity between $(W_q T)$ and $(W_k V)^T$, applies softmax normalization, and aggregates $(W_v V)$ using the attention weights. In the multi-head case, the outputs of $H$ attention heads, each with independent projection parameters, are concatenated and mapped by $W_o \in \mathbb{R}^{D \times D}$:
\begin{equation}
    \text{MHA}_{T \rightarrow V}(T, V) = W_o[\text{head}_1; \ldots; \text{head}_H]
\end{equation}
Each $\text{head}_h$ is computed as in Eq.~(\ref{Eq_attention}).

One notices that this hierarchical information fusion mechanism and multiple residual connections are used, which are capable of mitigating gradient vanishing during training while exhibiting strong representational ability.


\subsection{Asymmetric Group Preference Optimization}
\label{subsec-agpo}

Addressing the twin challenges of high variance and low sample utilization in policy gradient methods for graph, preference optimization (PO) approaches such as POMO+PO \citep{pan2025preference} offer a promising alternative by replacing quantitative losses with qualitative comparisons through the Bradley-Terry (BT) model \citep{bradley1952rank}, thus establishing a preference-based mechanism for graph optimization. Despite this conceptual shift, the underlying on-policy REINFORCE framework constrains their sample efficiency. We propose a reinforcement learning strategy, AGPO, designed to maximize the benefits of BT-model-based preference learning. AGPO explicitly compares the relative quality within groups of solution trajectories. This formulation enables fundamentally more stable optimization and unlocks significantly higher learning efficiency and sample utilization compared to existing preference-based methods.

\begin{algorithm}[!htb]
\caption{The AGPO Training}
\label{algo:AGPO}
\KwIn{
    Training set $S$, number of starting nodes per sample $N$, number of total training steps $T$, batch size $B$ 
}
\KwOut{
    Optimized policy $\pi_{\theta}$ 
}
Initialize policy network $\pi_{\theta}$\;

\For{step $= 1$ \textbf{to} $T$}{
    Collect trajectories $\{\tau_{\text{ref}}^i\}_{i=1}^B$ using $\pi_{\theta}$\;
    
    Compute rewards $R(\tau_{\text{ref}}^i)$ and memories $M_{\text{ref}}^i$\;
    
    Select top-$k$ trajectories: $\{\tau_{\text{pref}}^i\} \gets \text{Top}_k(R(\tau_{\text{ref}}^i))$\;
    
    Select bottom-$k$ trajectories: $\{\tau_{\text{nonpref}}^i\} \gets \text{Bottom}_k(R(\tau_{\text{ref}}^i))$\;
    
    \For{iteration $= 1$ \textbf{to} $K$ }
    {
        Collect trajectories $\{\tau_{\text{curr}}^i\}_{i=1}^B$ using $\pi_{\theta}$\;

        Compute log-probability ratios: 
        $\Delta_{\text{pref}} = \log\prod\limits_i\frac{\pi_{\theta}(\tau_{\text{pref}})}{\pi_{\text{ref}}(\tau_{\text{pref}})}$, 
        $\Delta_{\text{nonpref}} = \log\prod\limits_j\frac{\pi_{\theta}(\tau_{\text{nonpref}})}{\pi_{\text{ref}}(\tau_{\text{nonpref}})}$\;
        
        Formulate the loss function: 
        $\mathcal{L}_{\text{AGPO}} = -\mathbb{E}\left[\log\sigma\left(\beta_w\Delta_{\text{pref}} - \beta_l\Delta_{\text{nonpref}}\right)\right]$\;
        
        Update parameters: $\theta \gets \theta -\alpha\nabla_{\theta}\mathcal{L_{\text{AGPO}}}$\;
    }
}
\KwRet{Optimized policy $\pi_{\theta}$}
\end{algorithm}

Given a sequence of starting nodes from $V$ that generate corresponding trajectories \( \tau' = (\tau^1, \tau^2, \ldots, \tau^n) \), and assuming these trajectories have rewards in descending order \( r_{\tau^1} > r_{\tau^2} > \cdots > r_{\tau^n} \), we can formally obtain preference pairs \( D(r_{\tau^i}, r_{\tau^j}) \) where \( i < j \). For all samples, the joint likelihood is:
\begin{equation}
J(\theta) = \prod_{i<j} P(r_{\tau^i} \succ r_{\tau^j} | V)
\end{equation}
To facilitate optimization, we take the logarithm of the joint likelihood and add a negative sign to obtain the loss function:
\begin{equation} 
\label{eq:loss}
    L(\theta) = -\sum_{i<j} \log P(r_{\tau^i} \succ r_{\tau^j} | V)
\end{equation} 
Substituting the BT formula as in PO, 
we get:
\begin{equation}
\begin{aligned}
L(\theta) &= -\sum_{i<j} \log \left( \frac{\exp(r_{\tau^i})}{\exp(r_{\tau^i}) + \exp(r_{\tau^j})} \right) \\
&= -\sum_{i<j} \left[ \log \sigma(r_{\tau^i} - r_{\tau^j}) \right]
\end{aligned}
\end{equation}
Following the approach in Direct Preference Optimization (DPO) \citep{rafailov2023direct}, we parameterize the reward function as the difference in log probabilities between the current policy and a reference policy. Here, $\pi_{\text{theta}}$ and  $\pi_\text{ref}$ indicate the optimized and reference parameters:
\begin{equation}
r(x, y) = \beta \log \frac{\pi_\theta(\tau | V)}{\pi_{\text{ref}}(\tau | V)}
\end{equation}
For routing, departing from the depot, not every starting node leads to optimal solutions, which is particularly evident in CVRP where the quality of solutions can vary significantly depending on the starting node. To address this challenge within the multi-start nodes, we introduce an innovative grouping strategy to improve robustness.

Rather than considering preferences between individual trajectories, which can be unstable during training, we partition the trajectories into high-quality and low-quality groups. Specifically, we unify the probabilities of the top $k$ trajectories $\prod_{i=1}^{k} \pi_{\theta}(\tau^i)$ as a group optimal probability, and similarly consolidate the probabilities of the bottom $k$ trajectories $\prod_{j=n-k+1}^{n} \pi_{\theta}(\tau^j)$ as a group suboptimal probability. This grouping approach creates more robust probability pairs, reducing sensitivity to individual trajectory variations and improving training stability. The reward function for our grouped trajectories becomes:
\begin{equation}
r(x, y) = \beta \log {\prod_i\frac{\pi_\theta(\tau^i | V)}{\pi_{\text{ref}}(\tau^i | V)}}
\end{equation}

We further establish asymmetric coefficients ($\beta_w > \beta_l$) to place greater emphasis on improving the quality of positive samples. Here, $\tau_{\text{pref}}$ and $\tau_{\text{nonpref}}$ indicate the groups of highest and lowest reward trajectories, respectively:
\begin{equation}
r_w(x, y) = \beta_w \log \frac{\pi_\theta(\tau_{\text{pref}} | V)}{\pi_{\text{ref}}(\tau_{\text{pref}} | V)}
\end{equation}
\begin{equation}
r_l(x, y) = \beta_l \log \frac{\pi_\theta(\tau_{\text{nonpref}} | V)}{\pi_{\text{ref}}(\tau_{\text{nonpref}} | V)}
\end{equation}

This asymmetric weighting allows the model to focus more on improving promising solutions while maintaining a suitable learning signal from suboptimal trajectories. The training process of AGPO is presented in Algorithm \ref{algo:AGPO}. It begins with policy initialization, followed by iterative collection of reference trajectories, identification of preferred and non-preferred groups, and policy updates through the asymmetric preference optimization approach.

\begin{figure}[!t]
    \centering
    \includegraphics[width=\linewidth]{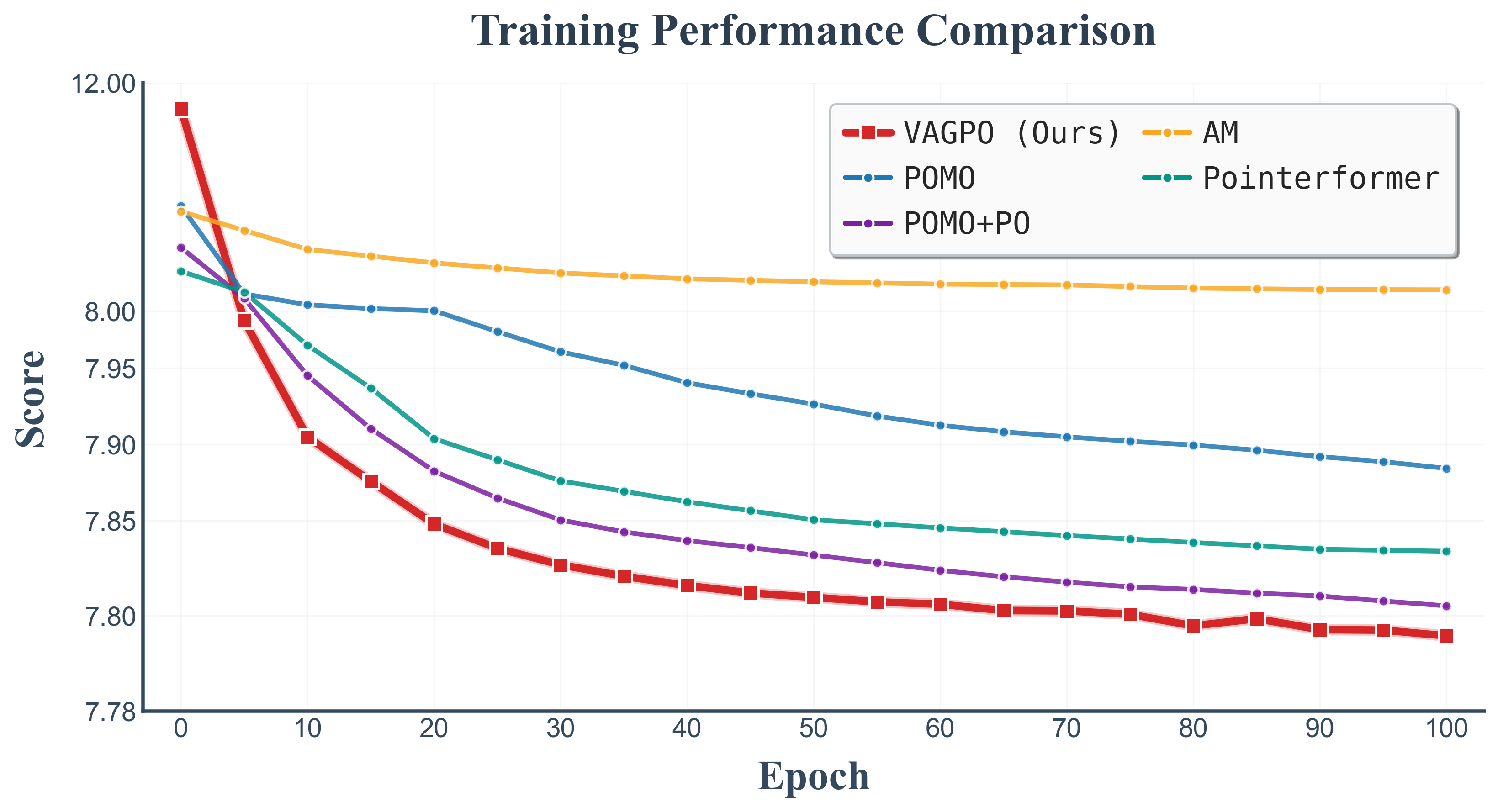}
    \caption{Training performance among different algorithms}
    \label{fig:tsp_curve}
\end{figure}

\begin{table*}[!t]
\centering
\caption{Experimental results on TSP instances}
\label{tab:tsp_results}
\resizebox{1.55\columnwidth}{!}{
\begin{tabular}{l l ccc ccc}
\toprule
\multirow{2}{*}{Method}&\multirow{2}{*}{Type} & \multicolumn{3}{c}{TSP-50} & \multicolumn{3}{c}{TSP-100} \\
\cmidrule(lr){3-5} \cmidrule(lr){6-8}
& & Len. & Gap &Time& Len. & Gap & Time \\
\midrule
Concorde       &Exact & 5.69  & 0.00\% &  (13m) & 7.76 & 0.00\% & (1h)  \\
LKH3          &OR & 5.69 & 0.00\% & (6m)  & 7.76 & 0.00\% &  (25m) \\
Gurobi        &OR  & 5.69 & 0.00\%  & (2m)  & 7.76 & 0.00\%  & (17m) \\
OR Tools      &OR & 5.85 & 2.87\%  & (5m)  & 8.06 & 3.86\%  & (23m) \\
POMO, no augment. &RL+AS & 5.70 & 0.12\% & (2s) & 7.80 & 0.46\% & (11s) \\
POMO, $\times$8 augment. &RL+AS & 5.69& 0.03\% &  (16s) &7.77 & 0.15\%  &  (1.0m)\\
Pointerformer & RL & 5.69 & 0.02\%  & (12s)  & 7.77 & 0.15\%  & (1m)  \\
DIFUSCO &SL+G   & 5.71 & 0.45\%  & (9m)  & 7.85 & 1.21\%  & (9m)  \\
DIFUSCO &SL+S  & 5.69 & 0.09\%  & -  & 7.78 & 0.23\%  & -  \\
POMO+PO, $\times$8 augment. &PO+AS & 5.69& 0.02\% &  (16s) &7.76 & 0.07\%  &  (1.0m)\\
\hdashline
\textbf{VAGPO, no augment.} &PO+AS&   5.69   &    0.05\%    &  (13s) &    7.78    &  0.25\%  &  (17s)    \\
\textbf{VAGPO, $\times$8 augment.} & PO+AS&  5.69    &  \textbf{0.01\%}  &(1.0m) &     7.76 &  \textbf{0.03\%} &  (2.2m)    \\
\bottomrule
\end{tabular}
}
\end{table*}

\begin{table*}[!htb]
\centering
\caption{Generalization results on TSP instances where POMO and VAGPO use $\times$8 augmentation. "Size $N$" refers to the corresponding models trained on the instances with $N$ nodes}
\label{tab:generalization}
\resizebox{2\columnwidth}{!}{
\begin{tabular}{l ccc ccc ccc ccc}
\toprule
\multirow{2}{*}{Method} & \multicolumn{3}{c}{TSP-50}& \multicolumn{3}{c}{TSP-100} & \multicolumn{3}{c}{TSP-500}&\multicolumn{3}{c}{TSP-1000} \\
\cmidrule(lr){2-4}\cmidrule(lr){5-7} \cmidrule(lr){8-10}\cmidrule(lr){11-13}
& Len. & Gap  & Time & Len. & Gap & Time& Len. & Gap & Time& Len. & Gap & Time \\
\midrule
Concorde   & 5.69 & 0.00\% & (13m)  & 7.76 & 0.00\% &(1h) &16.55&0.00\%&(4h)& 23.12 & 0.00\% & (7h) \\
LKH3   & 5.69 & 0.00\% & (6m)  & 7.76 & 0.00\% & (25m)&16.55&0.00\%&(46m)& 23.12 & 0.00\% & (3h) \\
H-TSP \citep{pan2023h} & - & - & - & - & - & - & - & -&- &24.65& \textbf{6.62\%}& (0.33s) \\
\hdashline
POMO, size 50 & 5.69 & 0.03\% & (16s) & 7.81 & 0.69\% & (1.0m) & 21.65 & 30.82\%&(1.5h) &33.52&44.98\%& (11.6h) \\
POMO, size 100 & 5.69 & 0.07\% & (16s) & 7.77 & 0.14\% & (1.0m) & 19.49 & 17.76\%&(1.5h)&30.71&32.82\%& (11.6h) \\


\hdashline
\textbf{VAGPO, size 50} &  5.69  &  \textbf{0.01\%} &  (1.0m)  &  7.78  & 0.30\% & (2.2m) & 22.47&35.77\%&(2.8h)  &33.95&46.86\%& (21.8h) \\
\textbf{VAGPO, size 100} &5.69 & 0.02\%& (1.0m) &7.76&  \textbf{0.03\%} & (2.2m)&  18.71  & \textbf{13.09\%} & (2.8h)&28.92& 25.08\% & (21.8h) \\

\bottomrule
\end{tabular}
}
\end{table*}

\section{EXPERIMENTAL RESULTS}
\label{sec-exp}
\subsection{Experimental Setting}
To comprehensively evaluate VAGPO, we use both randomly generated synthetic instances and established real-world benchmarks.

\textit{Random Generated Instances} \quad For training and evaluation, we generate TSP and CVRP instances on the fly. For all instances, node coordinates are uniformly sampled from the unit square $[0, 1]^2$. For TSP, we train models on instances with 50 and 100 nodes (TSP-50, TSP-100) and evaluate their generalization capabilities on larger instances with 500 and 1000 nodes (TSP-500, TSP-1000). For CVRP, we train on instances with 20, 50, and 100 customers (CVRP-20, CVRP-50, CVRP-100) and evaluate generalization on larger instances with 500 customers (CVRP-500). Following standard settings \citep{kool2018attention, kwon2020pomo}, customer demands are sampled from a discrete uniform distribution $\{1, 2, \dots, 9\}$, and vehicle capacities are set to 30, 40, 50, and 100 for CVRP-20, CVRP-50, CVRP-100, and CVRP-500, respectively. We generate 10,000 instances for each test set to ensure robust evaluation.

\textit{Real-world Instances} \quad To assess the model's performance on problems with non-uniform and structured node distributions, we use instances from the TSPLIB \citep{reinelt1991TSPLIB} and XML100 \citep{queiroga202110}, a widely recognized collection of benchmark problems for TSP and CVRP. We evaluate our model, trained on randomly generated 100 nodes instances, on a set of 29 TSPLIB instances with sizes ranging from 50 to 200 nodes, or a group of XML100 instances with sizes of 100 customers, without any retraining or fine-tuning. 

We compare our VAGPO against a number of leading baselines, including the exact solver Concorde \citep{applegate2006concorde}, the heuristic algorithm LKH3 \citep{helsgaun2017extension}, and several deep reinforcement learning models such as AM \citep{kool2018attention}, POMO \citep{kwon2020pomo}, and Pointerformer \citep{jin2023pointerformer}. We also include recent SOTA algorithms like DIFUSCO \citep{sun2023difusco} and POMO+PO \citep{pan2025preference}.

For all experiments, training used a single RTX 4090 GPU with a batch size of 64. Adam optimizer is used with a learning rate $\eta = 10^{-4}$ and a weight decay $w = 10^{-6}$. For AGPO, we set $\beta_w = 0.5$, $\beta_l = 0.1$, $\text{Top}_k = 10$. 

To strike a balance between efficiency and effectiveness in spatial representation learning, we chose ResNet-18 \citep{he2016deep} as our lightweight backbone, instead of more complex alternatives such as ResNet-50, Vision Transformer (ViT) \citep{dosovitskiy2020image}, or GNNs \citep{scarselli2008graph}. An epoch is defined as training across 100,000 randomly generated instances. Despite introducing only minor computational overhead, our method completes a full training epoch in under 13 minutes. For reference, under the same experimental conditions, POMO requires 5 minutes per epoch, and POMO+PO takes 9 minutes per epoch.

We follow the convention and report the time for solving 10,000 random instances of each problem. Similar to POMO, we have also performed inferences with $\times$8 instance augmentation, which has long been recognized as an effective post-processing technique. The model demonstrates robustness to variations in image resolution, with only negligible performance degradation observed when the resolution is adjusted during inference. 

\begin{table*}[!htb]
\centering
\caption{Experimental results on CVRP instances}
\label{tab:cvrp_results}
\resizebox{2\columnwidth}{!}{
\begin{tabular}{l ccc ccc ccc ccc}
\toprule
\multirow{2}{*}{Method} & \multicolumn{3}{c}{CVRP-20}& \multicolumn{3}{c}{CVRP-50} & \multicolumn{3}{c}{CVRP-100} & \multicolumn{3}{c}{CVRP-500} \\
\cmidrule(lr){2-4}\cmidrule(lr){5-7} \cmidrule(lr){8-10} \cmidrule(lr){11-13}
& Len. & Gap  & Time & Len. & Gap & Time& Len. & Gap & Time & Len. & Gap & Time \\
\midrule
LKH3           & 6.12 & 0.00\% & (2h)  & 10.38 & 0.00\% & (7h)& 15.68 & 0.00\% & (12h) & 40.06 & 0.00\% & (48h) \\
OR Tools       & 6.42 & 4.84\% & (2m) & 11.22 & 8.12\% & (12m) & 17.14&9.34\%&(1h) & - & - & - \\

POMO, no augment. & 6.17 & 0.82\% & (1s) & 10.49 & 1.14\% & (4s) & 15.83 & 0.98\%&(9s) & 48.34 & 20.64\% & (13.1m) \\

POMO, $\times$8 augment. &  6.14 &    0.21\%   &  (5s)     &   10.42   & 0.45\%     &  (26s)     &  15.73    &  0.32\%&(1.2m) & 44.81 & 11.85\% & (1.8h) \\
POMO+PO, $\times$8 augment. &  6.13 &    0.09\%   &  (5s)     &   10.41   & \textbf{0.32\%}   &  (26s)     &  15.73    &  0.26\%&(1.2m) & - & - & - \\

\hdashline

\textbf{VAGPO, no augment.} &   6.14   &  0.32\% &  (3s)    &    10.48   & 0.96\%  & (5s) &15.81 &0.82\% &(11s) & 46.88 & 17.02\% & (15.8m) \\

\textbf{VAGPO, $\times$8 augment.}  &  6.12  &  \textbf{0.05\%}  &  (23s)   & 10.42 & 0.38\% &(35s)& 15.72&\textbf{0.25\%}& (1.5m) & 44.25 & \textbf{10.46\%} & (2.1h) \\

\bottomrule
\end{tabular}
}
\end{table*}

\begin{table}[!htb]
\centering
\caption{Experimental results on real-world TSPLIB instances (50-200 nodes) where POMO and VAGPO use ×8 augmentation}
\label{tab:TSPLIB_small}
\resizebox{\columnwidth}{!}{
\begin{tabular}{l c ccc ccccc}
\toprule
\multirow{2}{*}{Instance} & \multirow{2}{*}{Opt.} &
\multicolumn{3}{c}{POMO} & \multicolumn{3}{c}{VAGPO} \\
\cmidrule(lr){3-5}\cmidrule(lr){6-8}
 &  & Len. & Gap (\%) & Time (s) & Len. & Gap (\%) & Time (s) \\
\midrule
eil51 &6.762&6.818& 0.829 & 0.278&\textbf{6.808} & \textbf{0.674} & 0.492 \\
berlin52 &4.398&4.399& 0.035 & 0.290 &\textbf{4.399}& \textbf{0.031} & 0.395 \\
st70 &6.818&6.840& 0.313 & 0.193 &\textbf{6.840}& \textbf{0.313} & 0.215 \\
eil76 &7.472&7.561& 1.184 & 0.218 &\textbf{7.561}& \textbf{1.184} & 0.273 \\
pr76 &5.518&5.518& 0.000 & 0.206 &5.520& 0.023 & 0.340 \\
rat99 &5.685&5.793& 1.898 & 0.240 &\textbf{5.727}& \textbf{0.732} & 0.295 \\
kroA100 &5.407&5.429& 0.413 & 0.215 &\textbf{5.425}& \textbf{0.335} & 0.252 \\
kroB100 &5.627&5.645& 0.323 & 0.234 &5.665& 0.674 & 0.263 \\
kroC100 &5.286&5.296& 0.183 & 0.210 &\textbf{5.291}& \textbf{0.080} & 0.238 \\
kroD100 &5.461&5.507& 0.842 & 0.245 &\textbf{5.479}& \textbf{0.318} & 0.347 \\
kroE100 &5.624&5.649& 0.450 & 0.217 &\textbf{5.639}& \textbf{0.269} & 0.307 \\
rd100 &8.065&8.065& 0.005 & 0.244 &\textbf{8.065}& \textbf{0.005} & 0.344 \\
eil101 &8.500&8.657& 1.844 & 0.214 &8.673& 2.037 & 0.355 \\
lin105 &4.755&4.780& 0.521 & 0.214 &4.789& 0.718 & 0.316 \\
pr107 &5.370&5.404& 0.637 & 0.235 &\textbf{5.392}& \textbf{0.402} & 0.243 \\
pr124 &6.263&6.286& 0.365 & 0.222 &\textbf{6.268}& \textbf{0.076} & 0.273 \\
bier127 &6.937&7.545& 8.774 & 0.237 &\textbf{7.060}& \textbf{1.782} & 0.240 \\
ch130 &8.831&8.845& 0.157 & 0.226 &\textbf{8.841}& \textbf{0.105} & 0.307 \\
pr136 &8.545&8.611& 0.774 & 0.252 &\textbf{8.645}& \textbf{0.697} & 0.275 \\
pr144 &5.358&5.382& 0.453 & 0.245 &\textbf{5.364}& \textbf{0.106} & 0.273 \\
ch150 &9.337&9.387& 0.527 & 0.274 &9.396& 0.624 & 0.280 \\
kroA150 &6.710&6.757& 0.696 & 0.243 &6.788& 1.158 & 0.320 \\
kroB150 &6.657&6.735& 1.167 & 0.239 &\textbf{6.734}& \textbf{1.153} & 0.278 \\
pr152 &5.368&5.435& 1.240 & 0.250 &\textbf{5.397}& \textbf{0.525} & 0.319 \\
u159 &8.092&8.169& 0.951 & 0.267 &\textbf{8.168}& \textbf{0.936} & 0.291 \\
rat195 &8.038&8.693& 8.150 & 0.289 &\textbf{8.321}& \textbf{3.524} & 0.410 \\
d198 &3.917&4.595& 17.290 & 0.305 &\textbf{4.500}& \textbf{14.868} & 0.414 \\
kroA200 &7.452&7.569& 1.577 & 0.321 &\textbf{7.569}& \textbf{1.565} & 0.489 \\
kroB200 &7.466&7.577& 1.493 & 0.304 &\textbf{7.552}& \textbf{1.160} & 0.415 \\
\midrule
Mean &6.542&6.653& 1.831 & 0.246 &\textbf{6.616} & \textbf{1.244} & 0.319 \\
\bottomrule
\end{tabular}
}
\end{table}

\begin{table}[!htb]
\centering
\caption{Experimental results on real-world CVRPLIB instances (100 customers) where POMO and VAGPO use ×8 augmentation}
\label{tab:CVRPLIB_100}
\resizebox{\columnwidth}{!}{
\begin{tabular}{l c ccc ccc}
\toprule
\multirow{2}{*}{Instance} & \multirow{2}{*}{Opt.} & \multicolumn{3}{c}{POMO} & \multicolumn{3}{c}{VAGPO} \\
\cmidrule(lr){3-5}\cmidrule(lr){6-8}
& & Len. & Gap (\%) & Time (s) & Len. & Gap (\%) & Time (s) \\
\midrule
G1111 & 34.662 & 36.245 & 4.767 & 0.243 & \textbf{35.027} & \textbf{1.063} & 0.366 \\
G1122 & 21.392 & 21.898 & 2.302 & 0.224 & \textbf{21.806} & \textbf{1.888} & 0.347 \\
G1133 & 15.869 & 16.093 & 1.405 & 0.200 & \textbf{16.065} & \textbf{1.224} & 0.337 \\
G1144 & 13.031 & 13.312 & 2.158 & 0.195 & \textbf{13.249} & \textbf{1.679} & 0.341 \\
G1214 & 12.492 & 13.360 & 7.239 & 0.200 & \textbf{12.650} & \textbf{1.304} & 0.321 \\
G1225 & 9.147 & 9.679 & 5.816 & 0.203 & \textbf{9.454} & \textbf{3.334} & 0.345 \\
G1236 & 7.155 & 8.323 & 16.744 & 0.200 & \textbf{7.500} & \textbf{4.930} & 0.323 \\
G1255 & 9.228 & 9.936 & 7.845 & 0.199 & \textbf{9.451} & \textbf{2.462} & 0.323 \\
G1311 & 37.074 & 38.869 & 5.083 & 0.232 & \textbf{37.417} & \textbf{0.938} & 0.347 \\
G1322 & 22.048 & 22.591 & 2.452 & 0.236 & \textbf{22.501} & \textbf{2.047} & 0.355 \\
G1333 & 16.916 & 17.223 & 1.798 & 0.222 & \textbf{17.190} & \textbf{1.592} & 0.337 \\
G1366 & 8.400 & 9.244 & 10.114 & 0.206 & \textbf{8.877} & \textbf{5.726} & 0.312 \\
G2114 & 11.191 & 12.012 & 7.373 & 0.201 & \textbf{11.321} & \textbf{1.165} & 0.322 \\
G2125 & 9.666 & 10.084 & 4.352 & 0.198 & \textbf{9.840} & \textbf{1.787} & 0.332 \\
G2136 & 8.450 & 9.444 & 11.837 & 0.214 & \textbf{8.825} & \textbf{4.449} & 0.332 \\
G2141 & 24.166 & 26.057 & 7.615 & 0.249 & \textbf{25.154} & \textbf{4.050} & 0.347 \\
G2211 & 28.815 & 30.499 & 6.262 & 0.241 & \textbf{29.196} & \textbf{1.373} & 0.357 \\
G2222 & 16.246 & 16.736 & 3.013 & 0.227 & \textbf{16.627} & \textbf{2.376} & 0.339 \\
G2233 & 10.765 & 11.033 & 2.501 & 0.202 & \textbf{10.968} & \textbf{1.891} & 0.331 \\
G2252 & 16.326 & 16.969 & 3.916 & 0.210 & \textbf{16.717} & \textbf{2.393} & 0.349 \\
G2314 & 10.750 & 11.695 & 8.854 & 0.209 & \textbf{10.885} & \textbf{1.248} & 0.336 \\
G2325 & 9.141 & 9.538 & 4.342 & 0.196 & \textbf{9.392} & \textbf{2.749} & 0.330 \\
G2336 & 7.920 & 8.994 & 13.540 & 0.207 & \textbf{8.237} & \textbf{4.003} & 0.318 \\
G2363 & 12.705 & 12.989 & 2.215 & 0.204 & \textbf{12.946} & \textbf{1.873} & 0.326 \\
G3111 & 50.731 & 52.703 & 3.912 & 0.237 & \textbf{51.216} & \textbf{0.935} & 0.353 \\
G3122 & 27.989 & 28.563 & 2.007 & 0.220 & \textbf{28.458} & \textbf{1.664} & 0.336 \\
G3133 & 20.205 & 20.520 & 1.570 & 0.207 & \textbf{20.479} & \textbf{1.363} & 0.337 \\
G3144 & 16.034 & 16.367 & 2.088 & 0.197 & \textbf{16.292} & \textbf{1.611} & 0.320 \\
G3214 & 15.502 & 16.519 & 6.661 & 0.215 & \textbf{15.706} & \textbf{1.346} & 0.321 \\
G3225 & 11.344 & 11.971 & 5.712 & 0.203 & \textbf{11.644} & \textbf{2.760} & 0.329 \\
G3236 & 8.364 & 9.681 & 16.083 & 0.211 & \textbf{8.794} & \textbf{5.250} & 0.314 \\
G3255 & 11.705 & 12.575 & 7.633 & 0.206 & \textbf{12.025} & \textbf{2.823} & 0.314 \\
G3311 & 48.094 & 50.098 & 4.245 & 0.228 & \textbf{48.545} & \textbf{0.930} & 0.335 \\
G3322 & 28.212 & 28.847 & 2.193 & 0.214 & \textbf{28.698} & \textbf{1.699} & 0.330 \\
G3333 & 20.800 & 21.138 & 1.606 & 0.208 & \textbf{21.095} & \textbf{1.390} & 0.328 \\
G3366 & 9.671 & 10.501 & 8.749 & 0.205 & \textbf{10.087} & \textbf{4.385} & 0.315 \\
\midrule
Mean & 17.468 & 18.392 & 5.948 & 0.213 & \textbf{17.963} & \textbf{2.863} & 0.336 \\
\bottomrule
\end{tabular}
}
\end{table}

\subsection{Comparative Results on Generated Instances}
We conduct training and evaluation on TSP instances with 50 and 100 nodes, and assess generalization by testing the trained models on larger instances with 500 and 1000 nodes. Table~\ref{tab:tsp_results} presents the quantitative results, comparing our method with existing baselines. Performance is evaluated using three metrics: length (total tour cost), gap (optimality gap w.r.t. Concorde), and time (inference efficiency). The optimal solutions for the test sets are computed using Concorde.

As shown in Table~\ref{tab:tsp_results}, VAGPO consistently outperforms POMO and POMO+PO, achieving smaller optimality gaps with only a modest increase in inference time. Compared to SOTA methods such as DIFUSCO, VAGPO offers a superior solution quality while maintaining lower inference latency. Moreover, our approach exhibits the shortest training time among all baselines, highlighting its strong trade-off between learning efficiency and solution accuracy.

To assess generalization, we evaluate models trained on TSP-50 and TSP-100 directly on larger instances. As shown in Table~\ref{tab:generalization}, VAGPO maintains competitive solution quality and consistently outperforms POMO. This demonstrates that our approach combines low training cost with efficient inference, enabling scalable optimization without significant computational overhead.


Figure~\ref{fig:tsp_curve} shows the training curves of various methods on TSP-100, highlighting the efficiency and stability of our learning process. Notably, VAGPO converges to competitive performance within just one-tenth of the training epochs required by POMO, demonstrating significantly improved learning efficiency. While some baselines achieve marginally better final results, they do so at substantially higher training costs. In contrast, VAGPO offers excellent cost-effectiveness, making it particularly well-suited for resource-constrained and latency-sensitive applications.


We further evaluate the effectiveness of our model on CVRP across various scales. As reported in Table~\ref{tab:cvrp_results}, VAGPO achieves near-optimal results for instances with 20, 50, and 100 nodes, with performance measured relative to the leading LKH3. Note that the results of 500 nodes are obtained using models trained on 100-node instances to test generalization. Notably, no known algorithm can solve 10,000-instance CVRP sets to optimality within a reasonable time-frame. In total, our approach matches or exceeds the performance of recent SOTA methods, offering competitive solution quality and computational efficiency in both training and inference.

\begin{table}[!t]
\centering
\caption{Ablation study configurations on TSP-100. Variants 1–4 test a single component. Variants 5–6 examine different vision encoder settings. Variants 7–9 explore various AGPO parameter configurations}
\label{tab:ablation_study}
\begin{tabular}{lccccc}
\toprule
\multirow{2}{*}{Exp.} & \multirow{2}{*}{\makecell{Backbone}} & \multirow{2}{*}{Frozen} & \multicolumn{3}{c}{AGPO Params} \\
\cmidrule(lr){4-6}
& & & $\beta_w$ & $\beta_l$ & $\text{Top}_k$ \\
\midrule
Variant 1 & \ding{55} & N/A & 0.5 & 0.1 & 10\\ 
Variant 2 & ViT & \ding{52} & N/A  &N/A & N/A \\
Variant 3 & GNNs & \ding{52} & N/A  &N/A & N/A \\
Variant 4 & ResNet-18 & \ding{52} & N/A  &N/A & N/A \\
\hdashline
Variant 5 & ResNet-18 & \ding{55} & 0.5 & 0.1 & 10\\
Variant 6 & ResNet-50 & \ding{52} & 0.5 & 0.1 & 10\\
\hdashline
Variant 7 & ResNet-18 & \ding{52} & 0.5 & 0.5 & 10\\
Variant 8 & ResNet-18 & \ding{52} & 0.5 & 0.1 & 1\\
Variant 9 & ResNet-18 & \ding{52} & 0.5 & 0.1 & 50\\
\hdashline
\textbf{VAGPO} & ResNet-18 & \ding{52} & 0.5 & 0.1 & 10\\
\bottomrule
\end{tabular}
\end{table}

\subsection{Comparative Results on Real-world Instances}
To evaluate the generalization capability of the proposed VAGPO, we conduct experiments on well-established real-world benchmark instances for both TSP and CVRP. In all experiments, node coordinates are normalized to the unit square $[0,1]^2$, and solution quality is measured by the deviation of the obtained route length from the known optimal value. Both the optimality gap and runtime are reported for each algorithm to ensure a fair comparison. As a constructive method, VAGPO is compared against the SOTA constructive baseline POMO.

For TSP, we use instances from TSPLIB with node counts ranging from 50 to 200. For CVRP, we test on the XML100 subset of CVRPLIB, where each instance involves 100 customers. Coordinates are normalized, and customer demands are scaled relative to vehicle capacity. To ensure systematic and comprehensive evaluation, a representative subset of instances is selected to span four major characteristics: depot location (random, centered, or cornered), customer distribution (random, clustered, or random-clustered), demand pattern (seven types with varying mean and variance), and route length (six levels from very short to ultra-long). Instances are organized by sequentially varying these four dimensions to cover both typical and extreme configurations while avoiding redundancy. From each configuration, the first instance is selected, resulting in 36 representative groups in total. Each group is denoted as G+ABCD, where the four digits encode the specific combination of these dimensions, as shown in Table~\ref{tab:CVRPLIB_100}.

The experimental results, summarized in Table~\ref{tab:TSPLIB_small} for TSP and Table~\ref{tab:CVRPLIB_100} for CVRP, show that VAGPO consistently outperforms POMO across most instances, highlighting its robustness and strong generalization ability to real-world cases with diverse node distributions and varying problem scales.


\section{ABLATION STUDY}
\begin{figure}[!t]
    \centering
    \includegraphics[width=1\linewidth]{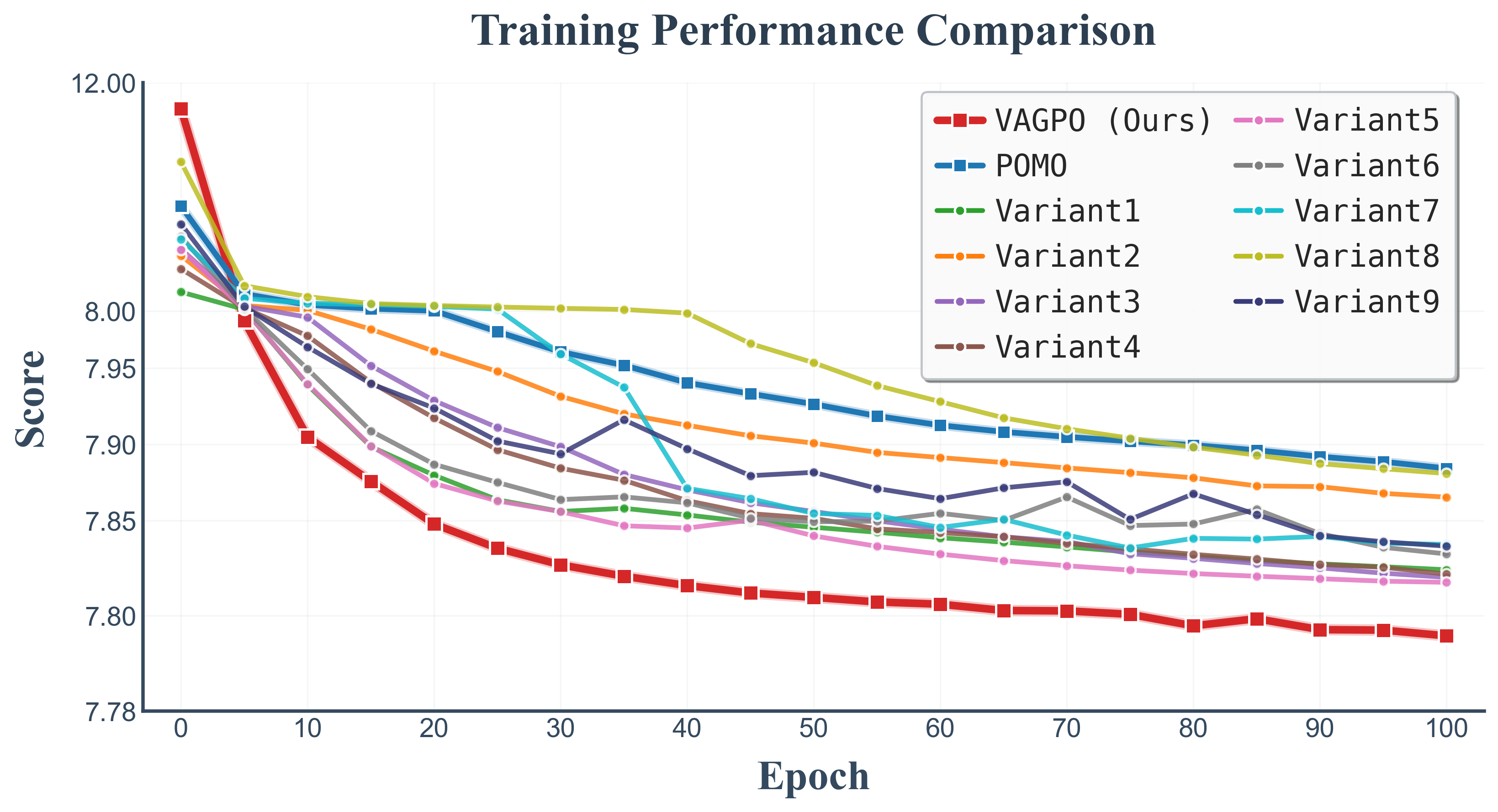}
    \caption{Training performance on VAGPO and variants}
    \label{fig:ablation}
\end{figure}

We perform ablation studies to quantify the impact of major design choices and validate the effectiveness of individual components in Table~\ref{tab:ablation_study} and visualized in Figure~\ref{fig:ablation}.
The training curves highlight the cumulative benefits of our architectural components, with the full model achieving high-quality solutions approximately 30\% faster than all ablated variants. Each component demonstrably contributes to improved performance, yielding faster convergence and better solution quality across diverse instances. On TSP-100, the systematic evaluation reveals clear performance gains attributable to each design element, underscoring the overall effectiveness of our approach.

As shown in Table~\ref{tab:ablation_study}, ResNet achieves a better accuracy–efficiency trade-off than GNN and ViT backbones. While GNNs and ViTs provide redundant global information and train two to four times slower, ResNet preserves crucial local spatial details with lower cost. ResNet-18 performs best overall, capturing essential spatial relations from sparse inputs while maintaining high efficiency. Freezing pre-trained weights further stabilizes early training, confirming ResNet-18 as the most effective backbone choice.


Similarly, Table~\ref{tab:ablation_study} compares different AGPO configurations. The optimized asymmetric setup delivers consistent gains over the baseline by emphasizing higher-quality positive samples, which is particularly beneficial when the optimal starting nodes are uncertain. The Top-$k$ strategy further improves sample utilization and accelerates preference estimation. Results show that moderate $k$ values yield the best balance between efficiency and diversity, leading to faster convergence and shorter route lengths across test instances.

    

\section{CONCLUSION}

In this paper, we have introduced VAGPO, integrating visual representations with sequential processing for graph problems. By transforming graph instances into image-like formats and utilizing a cross-modal fusion mechanism, our approach effectively captures both spatial relationships and sequential dependencies. The proposed AGPO training strategy significantly improves convergence speed through asymmetric optimization parameters and group-based trajectory evaluation. Experimental results demonstrate that VAGPO achieves competitive performance on TSP and CVRP while requiring fewer training epochs, showing strong generalization to larger instances. This balance of solution quality and computational efficiency makes VAGPO particularly valuable for practical applications. Future work could explore more sophisticated visual encoders and extensions to other graph problems with inherent spatial structures.

\bibliographystyle{plainnat}
\bibliography{sample-base}
\end{document}